\documentclass[11pt]{article}
	
	\newcommand{\blind}{0}
	
    \makeatletter
    \renewcommand\section{\@startsection {section}{1}{\z@}%
                                       {-3.5ex \@plus -1ex \@minus -.2ex}%
                                       {2.3ex \@plus.2ex}%
                                       {\normalfont\fontfamily{phv}\fontsize{16}{19}\bfseries}}
    \renewcommand\subsection{\@startsection{subsection}{2}{\z@}%
                                         {-3.25ex\@plus -1ex \@minus -.2ex}%
                                         {1.5ex \@plus .2ex}%
                                         {\normalfont\fontfamily{phv}\fontsize{14}{17}\bfseries}}
    \renewcommand\subsubsection{\@startsection{subsubsection}{3}{\z@}%
                                        {-3.25ex\@plus -1ex \@minus -.2ex}%
                                         {1.5ex \@plus .2ex}%
                                         {\normalfont\normalsize\fontfamily{phv}\fontsize{14}{17}\selectfont}}
    \makeatother
	
	\usepackage{amsmath}
	\usepackage{graphicx}
	\usepackage{enumerate}
	\usepackage{natbib} 
	\usepackage{url} 
	\usepackage{amsmath}
\usepackage{graphicx}
\usepackage{natbib}
\usepackage{url} 

\usepackage{braket,amsfonts}

\usepackage{array}

\usepackage[caption=false]{subfig}
\usepackage{pgfplots}
\usepackage{tikz-cd}

\usepackage{booktabs}

\usepackage{amsopn}

\usepackage{algpseudocode}
\usepackage{algorithm}

\usepackage{graphicx,epstopdf}
	
\begin{document}
		
		\def\spacingset#1{\renewcommand{\baselinestretch}%
			{#1}\small\normalsize} \spacingset{1}
		
		\if0\blind
		{
			\title{\bf \emph{Projection Pursuit Gaussian Process Regression}}
			\author{Gecheng Chen and Rui Tuo \\
			Wm Michael Barnes '64 Department of Industrial and Systems Engineering, \\Texas A\&M University, College Station, U.S. }

			\date{}
			\maketitle
		} \fi
		
		\if1\blind
		{

            \title{\bf \emph{IISE Transactions} \LaTeX \ Template}

\bigskip
			\bigskip
			\bigskip
			\begin{center}
				{\LARGE\bf \emph{IISE Transactions} \LaTeX \ Template}
			\end{center}
			\medskip
		} \fi
		\bigskip
		
	\begin{abstract}
A primary goal of computer experiments is to reconstruct the function given by the computer code via scattered evaluations. Traditional isotropic Gaussian process models suffer from the curse of dimensionality, when the input dimension is relatively high given limited data points. Gaussian process models with additive correlation functions are scalable to dimensionality, but they are more restrictive as they only work for additive functions. In this work, we consider a projection pursuit model, in which the nonparametric part is driven by an additive Gaussian process regression. We choose the dimension of the additive function higher than the original input dimension, and call this strategy ``dimension expansion''. We show that dimension expansion can help approximate more complex functions. A gradient descent algorithm is proposed for model training based on the maximum likelihood estimation. Simulation studies show that the proposed method outperforms the traditional Gaussian process models. The Supplementary Materials are available online. 
	\end{abstract}
			
	\noindent%

	\spacingset{1.5} 

\section{Introduction}\label{sec:intro}

 Contemporary practices in engineering and physical sciences have made increasing use of (deterministic) computer simulations, in disciplines including aerospace designs, material science, and biomedical studies. One of the central research topics is to build an accurate surrogate model to emulate computer simulations. Gaussian process regression \citep{Rasmussen2006gaussian,santner2003design} is one of the most popular surrogate models. Various modifications and extensions of the standard Gaussian process regression models have been proposed to address the specific needs in practical situations. An incomplete list of these methods include composite Gaussian processes \citep{ba2012composite}, treed Gaussian processes \citep{gramacy2008bayesian}, non-stationary models \citep{heaton2017nonstationary}, transformed approximately additive Gaussian processes \citep{lin2019transformation}, etc.
    
Data analysis for computer simulations usually suffers from the ``small data'' issue, because the computer simulation runs can be highly costly. For example, each run of a typical computational fluid dynamics model for aerospace engineering takes a few days or even weeks to run \citep{mak2018efficient}. Many computer simulations also pose the \textit{curse of dimensionality} problem, in the sense that the input dimension is relatively high so that building an accurate surrogate model based on limited data points becomes more challenging. Classic approaches for dimension reduction in computer experiments include sensitivity analysis \citep{durrande2013anova,oakley2004probabilistic,saltelli2010variance}, ridge approximation \citep{hokanson2018data,glaws2020inverse,pinkus1997approximating}. Variable selection for Gaussian processes models is considered in \cite{linkletter2006variable}, \cite{gu2019jointly} and \cite{constantine2014active}. In Gaussian process regression, it is also known that some correlation structures perform better in high-dimensional scenarios \citep{stein1999interpolation}. Recently, additive Gaussian process models have received considerable attention \citep{deng2017additive,duvenaud2011additive,lebarbier2005detecting,delbridge2020randomly,durrande2012additive,tripathy2016gaussian}. Although these models are more scalable to the input dimension, their capability of model fitting is lower because these models can only reconstruct additive functions precisely.

    In this work, we propose a novel surrogate modeling technique based on the projection pursuit methodology \citep{friedman1981projection} and additive Gaussian process models. Gaussian process regression can provide prediction variance as opposed to projection pursuit (neural networks). Additionally, unlike the conventional estimation approaches for projection pursuit \citep{ferraty2013functional,gilboa2013scaling,li2016high}, we suggest choosing a large number of intermediate nodes to introduce more model flexibility. Then we use the maximum likelihood estimation to identify the model parameters. A gradient descent algorithm is proposed to search the maximum of the likelihood function. In this work, we also find an error bound of the prediction error for Gaussian process regression with additive Mat\'ern correlation functions. Our theoretical results show that the prediction error of additive Gaussian process models is much lower than that given by isotropic Gaussian process models for high-dimensional problems, provided that a design with nice projection properties, such as a Latin hypercube design, is adopted.
    
    This article is organized as follows. In Section \ref{Sec:review}, we review the background of Gaussian process regression with isotropic and additive Mat\'ern correlation functions. 
    In Section \ref{Sec:PPGPR}, we introduce the proposed methodology, called the projection pursuit Gaussian process regression. An algorithm of the proposed method is given at the end of Section \ref{Sec:PPGPR}. In Section \ref{Sec:simulation} and \ref{more studies}, we conduct simulation studies to demonstrate the use of the proposed method, and show that the proposed method outperforms some existing methods. In Section \ref{application}, we shows that the performance of the proposed method is satisfactory through a real-world application. Concluding remarks are made in Section \ref{sec:discussion}.

    \section{Review on Gaussian process regression}\label{Sec:review}
    
    In this section, we review a simple version of the Gaussian process emulation \citep{santner2003design}. Let $Z$ be a stationary Gaussian process on $\mathbb{R}^d$ with mean zero, variance $\sigma^2$, and correlation function $\Phi$. Given scattered evaluations $(x_1,Z(x_1)),\ldots,(x_n,Z(x_n))$, one can reconstruct $Z$ using its conditional expectation
    \begin{eqnarray}
    \hat{Z}(x):=\mathbb{E}(Z(x)|Z(x_1),\ldots,Z(x_n))=r^T(x)K^{-1}Y,\label{GPR}
    \end{eqnarray}
for $x\in\mathbb{R}^d$, where $r(x):=(\Phi(x-x_1),\ldots,\Phi(x-x_n))^T, K=(\Phi(x_j-x_k))_{j k}$ for $j=1,\dots,n$ and $k=1,\dots,n$, and $Y=(Z(x_1),\ldots,Z(x_n))^T$.
    
    \subsection{Curse of dimensionality in Gaussian process regression with isotropic Mat\'ern correlation}\label{Sec:2}
    
    Curse of dimensionality is one of the fundamental challenges in various high-dimensional statistical and machine learning problems. In this section, we review how the curse of dimensionality can affect the prediction performance of Gaussian process regression.

The prediction error of the Gaussian process regression is  $$Z(x)-\hat{Z}(x)=Z(x)-\mathbb{E}(Z(x)|Z(x_1),\ldots,Z(x_n)),$$ 
which is a function of $x$. \cite{tuo2020kriging} study the rate of convergence of the prediction error under different function norms, under the assumption that the Gaussian process has an 
isotropic Mat\'ern correlation function \citep{santner2003design}, defined as
\begin{eqnarray}\label{isoMatern}
\Phi(x;\nu,\phi)=\frac{1}{\Gamma(\nu)2^{\nu-1}}(2\sqrt{\nu}\phi\|x\|)^\nu K_\nu(2\sqrt{\nu}\phi\|x\|),
\end{eqnarray}
where $\nu>0$ is the \textit{smoothness parameter}, $K_\nu$ is the modified Bessel function of the second kind, $\phi>0$ is the scale parameter. 
    
    To explain the curse of dimensionality issue posed by the isotropic Mat\'ern correlation functions, we refer to Theorem 3.3 of \cite{tuo2020kriging}, which states a lower bound of the maximum of the prediction error of an isotropic Gaussian process. For simplicity, we consider the expected maximum prediction error. Suppose the input region of interest is $\Omega$, and then the expected maximum prediction error is $\mathbb{E}\sup_{x\in\Omega}|Z(x)-\hat{Z}(x)|$. Here the expectation is taken over the randomness of the Gaussian process $Z(\cdot)$. Theorem 3.3 of \cite{tuo2020kriging} implies
    \begin{eqnarray}\label{isotropiclower}
    \mathbb{E}\sup_{x\in\Omega}|Z(x)-\hat{Z}(x)|\geq C \sigma n^{-\nu/d}\sqrt{\log n},
    \end{eqnarray}
    for a constant $C$ independent of $n$, $\sigma$ and the choice of the experimental design.
    
    The lower bound in (\ref{isotropiclower}) shows that the uniform error of a Gaussian process regression predictor with an isotropic Mat\'ern correlation is no less than a multiple of $n^{-\nu/d}\sqrt{\log n}$. This rate grows dramatically as $d$ increases with a fixed $\nu$. Therefore, when a Gaussian process model with an isotropic Mat\'ern correlation is considered, its prediction suffers from the curse of dimensionality, in the sense that, for a high-dimensional problem, acquiring extra data points cannot improve the prediction accuracy as effectively as in lower-dimensional problems.
    
    In Gaussian process regression, the curse of dimensionality is inevitable if the underlying function is indeed a realization of a Gaussian process with isotropic Mat\'ern correlation. The reason behind this is that the reproducing kernel Hilbert spaces generated by these correlation functions are too large in high-dimensional circumstances. 
    Fortunately, in most real applications, we confront much ``simpler'' high-dimensional functions. These functions admit certain ``sparse representation'', and therefore, at least theoretically, can be recovered at a much higher rate of convergence. In Section \ref{sec:additive}, we examine a special and simple structure of this kind.

\subsection{Additive models: accuracy and limitations}
\label{sec:additive}

 A scalable Gaussian process regression approach proceeds by equipping an additive correlation function. Denote $x=(x_{(1)},\ldots,x_{(d)})$. We consider the following function:
    \begin{eqnarray}\label{additive}
    \Phi(x)=\frac{1}{d}\sum_{j=1}^d \Phi_1(x_{(j)}),
    \end{eqnarray}
    where $\Phi_1$ denotes a one-dimensional correlation function. It is easily seen that $\Phi$ is positive definite if $\Phi_1$ is positive definite. Thus one can consider Gaussian process models with correlation (\ref{additive}). This approach is called the additive Gaussian process regression \citep{deng2017additive,duvenaud2011additive,lebarbier2005detecting}. 

    Compared to isotropic models, additive models are much more scalable to the dimensionality. 
    It can be shown that the rate of convergence of the uniform error is independent of $d$. Specifically, if $\Phi_1$ is a Mat\'ern correlation function with smoothness $\nu$, the uniform prediction error in (\ref{isotropiclower}) can have a rate of convergence $O(n^{-\nu}\sqrt{\log n})$; see our theoretical results in the Supplementary Materials. 
    
\begin{figure}
        \centering
        \includegraphics[width=0.8\textwidth]{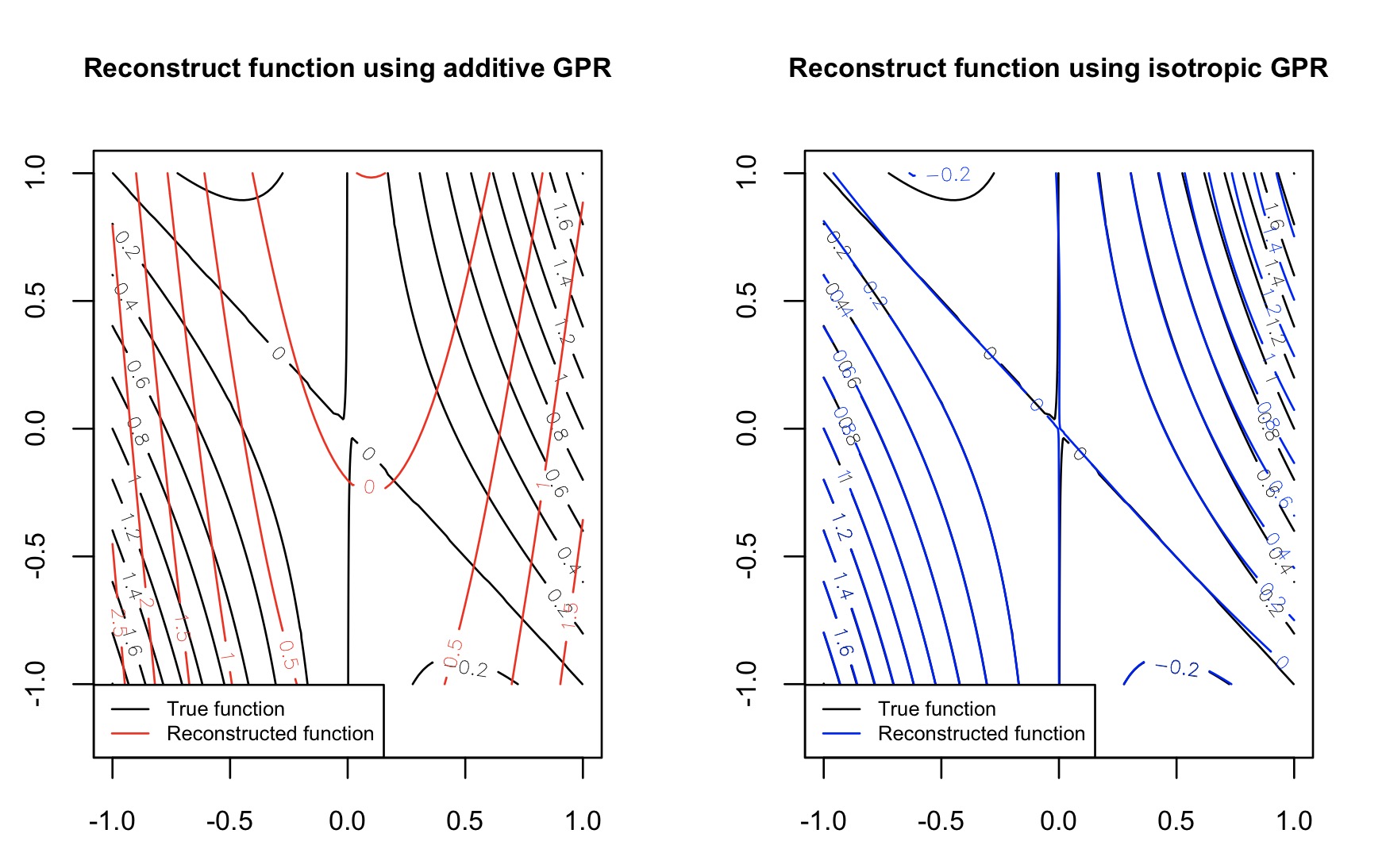}
        \caption{Contour plots of $f(x,y)=xy+x^2$ and the reconstructed functions by additive and isotropic Gaussian process regression (GPR) using a same 25-point random design between $-1$ and $1$. It can be seen that the isotropic model has a much better prediction performance.}
        \label{fig:additive_model}
    \end{figure}

    Despite the above advantages, the limitations of additive models are also evident. Only additive functions, i.e., the functions that can be decomposed as the sum of functions such that each of them relies on only one entry of $x$, can be accurately reconstructed. This assumption is \textit{not} true for most of the practical problems. Consider a two-dimensional input $(x,y)$. A simple non-additive function is $f(x,y)=xy+x^2$. Figure \ref{fig:additive_model} shows that the additive model cannot fit this function well, while the isotropic model works in this case.
    
\section{Projection pursuit Gaussian process regression}
\label{Sec:PPGPR}

    In this section, we propose a general approach to reconstruct multi-dimensional functions that admits more complicated sparse representations.
    To this end, we consider a model which is more flexible than additive Gaussian process models. Specifically, we employ the projection pursuit regression method \citep{friedman1981projection} to model the underlying function as
    
    \begin{eqnarray}\label{original additive structure}
        y(x)=f(w^T_1x,w^T_2 x,\ldots,w^T_M x),
    \end{eqnarray}
    where $w_1,\ldots,w_M$ are unknown vectors, $M$ is a positive integer, and $f$ is an additive function in the sense that $f$ can be written as
        \begin{eqnarray}
        f(w^T_1x,w^T_2 x,\ldots,w^T_M x)=f_1(w^T_1x)+f_2(w^T_2x)+\cdots+f_M(w^T_Mx),
    \end{eqnarray}
    with unknown univariate functions $f_1,\ldots,f_M$. 
    In other words, this model first applies a linear transformation on the input space, and then use an additive function to fit the responses. 
    
        
    
\begin{figure}[ht]
        \centering
        \includegraphics[width=0.8\textwidth]{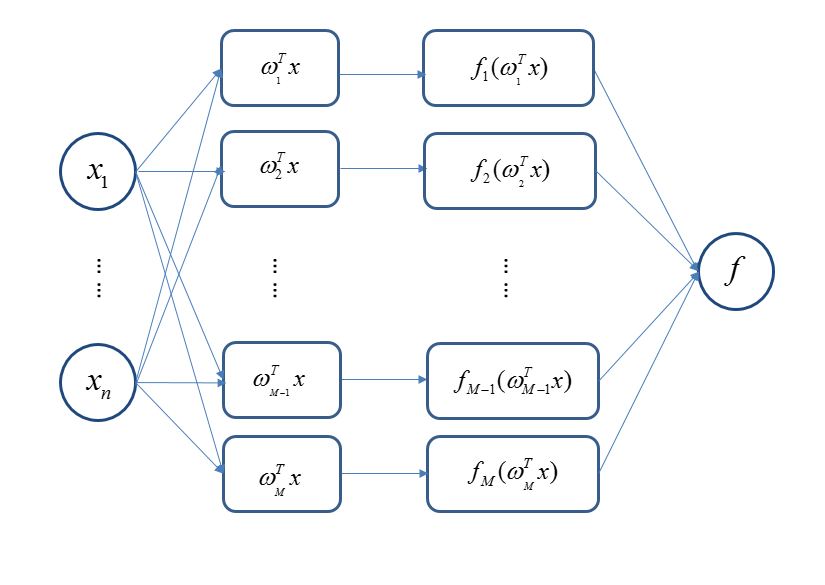}
        \caption{Network structure of PPGPR.}
        \label{fig:model structure}
    \end{figure}

    A projection pursuit model can be represented by a four-layer network shown in Figure \ref{fig:model structure}, which is similar to a neural network model. Neural networks have been widely used to enhance the precision of nonparametric regression \citep{goodfellow2016deep,hinton2006reducing,lecun2015deep,psichogios1992hybrid}; \cite{khoo2017solving} and \cite{tripathy2018deep} employ deep neural networks to reduce the dimension of data; \cite{wilson2011gaussian} combines neural networks with Gaussian process regression method to tackle multi-task problems. The main difference between the projection pursuit method and neural networks lies in the activation functions. In neural networks, the activation functions are chosen as fixed function, such as rectified linear unit (ReLU) functions. In contrast, the projection pursuit method uses estimated activation functions. In this work, we call the two hidden layers the \textit{transformation layers}.
    
        When $M=1$, the projection pursuit model reduces to a single index model, which provides a parsimonious way to implement multivariate non-parametric regression. By imposing suitable priors on the parameters, \cite{choi2011gaussian}, \cite{gramacy2012gaussian} and \cite{hu2013bayesian} use the Bayesian approach to estimate the parameters of the single index model. In \cite{wang2010estimation}, a dimension reduction method is applied to choose the number of nodes and then the link function is estimated using Gaussian process regression. In this work, we consider projection pursuit models with $M\gg 1$, which are much more flexible than single index models.
    
\begin{figure}[tbhp]
        \centering
        \subfloat[]{\label{fig:a}\includegraphics[width=0.45\linewidth]{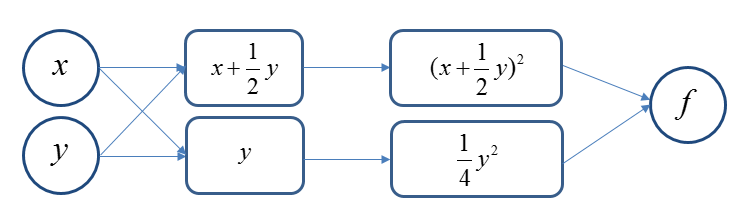}}
        \subfloat[]{\label{fig:b}\includegraphics[width=0.45\linewidth]{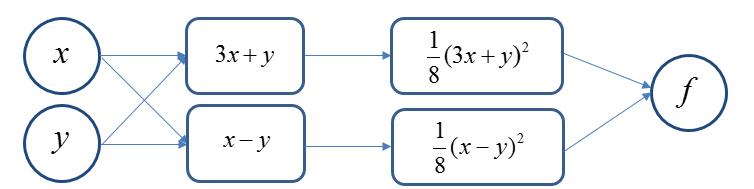}}
        \caption{Two different representations of $f(x,y)=xy+x^2$ via projection pursuit.}
        \label{fig:additive example}
    \end{figure}

     Given a sufficiently large $M$, it is known that the projection pursuit model can approximate any continuous function arbitrarily well \citep{friedman2001elements}.
    For example, the non-additive function $f(x,y)=xy+x^2$ can be represented by projection pursuit as shown in Figure  \ref{fig:additive example}. Figure \ref{fig:additive example} also shows that the representation is not unique.

    The non-uniqueness of the projected pursuit representation suggests that each of the ``directions'' $w_i$ may not be essential. In contrast, these vectors exhibit a ``synergistic effect'', so that they need to be estimated jointly. Consider the example shown in Figure \ref{fig:additive example} (a). Taking the direction $x+y/2$ along is not helpful in obtaining the underlying function $xy+x^2$; this direction makes sense only when it is paired by the direction $y$. This phenomenon differs from the classical results in linear models, in which the significant directions (usually defined by the principal components) are fixed, and their importance is ordered by the corresponding eigenvalues.
    
    Understanding this difference between the linear and nonlinear models helps build a better projection pursuit regression model. Traditionally, the projection pursuit method is usually regarded as a dimension reduction approach \citep{ferraty2013functional,gilboa2013scaling}, and greedy algorithms are usually applied to identify $w_i$'s \citep{gilboa2013scaling,muller2008functional,james2005functional}.
    These strategies have the following deficiencies: 1) it is often hard to accurately approximate the underlying functions through dimension reduction ($M\ll d$).  For example, the function $f(x,y)=xy+x^2$ cannot be recovered through a one-dimensional factor. 2) Greedy algorithms, which proceed by picking the current ``most significant'' direction in each step, cannot perform well when there is no order of importance in the directions, as in the example shown in Figure \ref{fig:additive example}. In this work, we propose a method, which conducts a dimension expansion ($M\geq  d$) to improve the approximation power substantially.
    
    When $M\geq d$, the projection pursuit model is in general non-identifiable; see Figure \ref{fig:additive example} for an example. The learning outcome on $w_i$'s are meaningless, and we only focus on the prediction of the underlying response at untried input points. Our numerical experience shows that as long as $M$ is large enough, the prediction performance of the proposed method is not heavily dependent on the specific value of $M$. We recommend choosing $M$ close to, but slightly less than the sample size $n$.

    In this work, we propose a novel approach, called the projection pursuit Gaussian process regression (PPGPR). 
    To reconstruct the underlying function, we need to: 1) estimate the weight parameters $w=(w_1,w_2,...,w_M)$; 2) reconstruct the combination function $f$ given $w$ using Gaussian process regression \citep{santner2003design,Rasmussen2006gaussian}.
    Recall that the design matrix is denoted as $X=(x_1,x_2,...,x_n)^T$, $x_i \in \mathbb{R}^d$ for $i=1,2,...,n$, and the response as $Y=(f(x_1),f(x_2),...,f(x_n))^T$.
    Now we employ the idea of Gaussian process regression to assume that $f$ is a realization of a Gaussian process. Specifically, we assume that the Gaussian process has mean zero and an additive correlation function (\ref{additive}). We believe that the mean zero assumption is not too restrictive because the model is already non-identifiable. 
    
    
    The training of the proposed method proceeds by an iterative approach.
    First, we choose an initial weight parameter $w$. Then we compute the initial correlation matrix $K_\omega=\left(\frac{1}{M}\sum_{k=1}^{M}\Phi(w_k^T(x_i-x_j))\right)_{ij}$ based on the initial $\omega$. Next, we invoke (\ref{GPR}) to reconstruct the underlying function $f$ as
    \begin{eqnarray}\label{prediction function}
        \hat f(x)=r^T(w^Tx)(K_\omega+\delta I)^{-1}Y,
    \end{eqnarray}
    where $\delta$ is a nugget term to enhance the numerical stability.
    
    Our goal is to seek for $w^*$ which maximizes the log-likelihood function of Gaussian Process Regression \citep{santner2003design}, that is,
    \begin{eqnarray}\label{objective function}
        \min_w(l(w))=\min_w(Y^T(K_w+\delta I)^{-1}Y+\log{\det(K_w+ \delta I)}).
    \end{eqnarray}
    We refer $l(w)$ to the \textit{model loss}. The gradient of $l(w)$ with respect to $w_k$ is
    \begin{eqnarray}\label{gradient}
        \frac{\partial l(w)}{\partial w_k}=-\frac{1}{M}\sum_{i=1}^{n}\sum_{j=1}^{n}(Y^T K_w^{-1}\frac{\partial K_w}{\partial w_k} K_w^{-1} Y+Tr(K_w^{-1}))(x_i-x_j)^T,
    \end{eqnarray}
    for $k=1,2,...,M$. The derivative of the matrix $K_w$ can be computed using the following facts. The derivative of the Mat\'ern correlation function is \citep{wendland2005scattered}
    $$    \frac{\partial }{\partial x}\Phi (x;\nu, \phi )=-\frac{2\nu \phi ^2 x}{\nu -1}\Phi\left(\sqrt{\frac{\nu }{\nu -1}}x;\nu -1,\phi \right).$$
    Then the gradient decent method can be applied here to find the minimizer via iteratively updating
        $$w_k\leftarrow w_k-\eta \frac{\partial l(w)}{\partial w_k},$$
    where $\eta$ is the step length for the gradient descent algorithm, and is referred to the \textit{learning rate} in the rest of this article. 

    When the algorithm converges or a stopping criterion is met, one can again reconstruct the underlying function using (\ref{prediction function}). Algorithm 1 lists the detailed steps of the proposed training method, each iteration (epoch) includes calculating the gradient for all weights and renewing the weights. To avoid overfitting, early-stopping criterion \citep{prechelt1998early} should be implemented when choosing $P$ (the number of epochs).

    Besides $P$, there are other hyper-parameters in the proposed methodology, including $M,\eta$ and the hyper-parameters of the covariance function. We refer the activity of adjusting these parameters to the \textit{tuning process}. Below is a list of our general recommendations for tuning.
    
    \begin{itemize}
        \item The proposed method does not use the ML estimators \citep{santner2003design} to estimate the hyper-parameters of the GP covariance because the ML estimators are likely to overfit with relatively small sample sizes \citep{santner2003design}. 
        \item Determining a proper learning rate $\eta$ through cross-validation such that it maintains a stable training process (i.e., the model loss decreases neither too sharply nor too slowly);
        \item Increasing the size of representation nodes $M$ until the performance on the testing points starts to deteriorate. In practice, we recommend considering $M$ in the range $[4d, 8d]$ in a $d$-dimensional problem;
        \item Adopting early stopping policies \citep{prechelt1998early} in the training process when choosing $P$ to avoid overfitting;
        \item Using cross-validation to choose the hyper-parameters of the covariance function.

    \end{itemize}

     More discussion regarding the tuning process is provided through a numerical study in Section \ref{Sec:tuning}.

    \begin{algorithm}[htb]
        \caption{Training steps for transformation weight $w$}
        \begin{algorithmic}[2]
            \Require
                design matrix $X=(x_1,x_2,...,x_n)$, response $Y=(y_1,y_2,...,y_n)$, initialized weight $w=(w_1,w_2,...,w_M)$, correlation function $\Phi$, learning rate $\eta$, number of iterations $P$
            \Ensure
                transformation weight $w$
            \For{$p$ in $1:P$}
                \State $X^{'}\leftarrow w^TX$
                \State $K_w \leftarrow \Phi(X^{'},X^{'})$
                \For{$k$ in $1:M$}
                    \State $grad_k \leftarrow -\frac{1}{M}\sum_{i=1}^{n}\sum_{j=1}^{n}(Y^T K_w^{-1}\frac{\partial K_w}{\partial w_k} K_w^{-1} Y+Tr(K_w^{-1}))(x_i-x_j)^T $
                    \State $w_k \leftarrow w_k - \eta \cdot grad_k$
                \EndFor
            \EndFor
        \end{algorithmic}
    \end{algorithm}

\section{Simulation Studies}
\label{Sec:simulation}

In this section, we examine the performance of the proposed method via simulation studies. Based on four numerical experiments, we will provide some guidelines for parameter tuning for PPGPR in Section \ref{Sec:tuning}. In Section \ref{Sec:numerical comparisons}, we compare the proposed method with some other prevailing algorithms and show the advantages of the proposed method.
    
    \subsection{Choice of tuning parameters}\label{Sec:tuning}
    In this section, we study how the choice of the hyper-parameters of PPGPR can affect its prediction performance. Recall that the hyper-parameters include the learning rate $\eta$, the size of nodes $M$ in the transformation layers, the number of epochs (iterations) $P$, the choice of the correlation function (Mat\'ern or Gaussian) and smoothness parameter $\nu$ if a Mat\'ern correlation is used. 
    
    In the rest of this subsection, we will use the Borehole function \citep{harper1983sensitivity} as the test function to study the performance of the proposed PPGPR under different choices of hyper-parameters. The Borehole function is defined as
    $$    y=\frac{2\pi T_u(H_u-H_l)}{\log(\frac{r}{r_w})[1+\frac{T_u}{T_l}+\frac{2LT_u}{\log(\frac{r}{r_w})r_w^2K_w}]},$$
    with the ranges for the eight variables given by $r_w \in(0.05,0.15)$, $r\in(
    100,50000)$, $T_u\in(63070,115600)$, $H_u\in(900,1110)$, $T_l\in(63.1,116)$, $H_l\in(700,820)$, $L\in(1120,1680)$ and $K_w\in(9855,12045)$. A Halton sequence\footnote{Halton sequence are deterministic low discrepancy sequences used to generate points in space for numerical experiments. The Halton sequences can be generated efficiently by the \texttt{R} package \texttt{SDraw}.} \citep{halton1964algorithm} with 40 samples are used as the training set inputs and 500 random samples are used as the testing set inputs. We consider different choices of the tuning parameters and compare the corresponding prediction performance in terms of the mean absolute percentage error (MAPE) \citep{makridakis1993accuracy}:
    \begin{eqnarray}\label{loss}
    MAPE=\frac{1}{n}\sum_{i=1}^{n}\left|  \frac{\hat{y}_i-y_i}{y_i} \right| ,
    \end{eqnarray}
    where $n=500$ is the size of testing samples; $\hat{y}$ and $y$ denote the predictive value and true value of a testing sample, respectively. 
    
    The details of the numerical experiments are described in Sections \ref{Sec:learning rate}-\ref{Sec:training epochs}. We choose $10^{-6}$ as the nugget term of (\ref{objective function}) in this section to avoid some numerical instability, see \cite{peng2014choice} for more guideline for choosing nugget terms. 
    
    \subsubsection{Learning rate $\eta$ and number of representation nodes $M$}\label{Sec:learning rate}
    In this experiment, a Mat\'ern correlation function with $\nu=2.5$ is used and training epochs $P=150$. We examine the performance of PPGPR under different learning rates and different node sizes in the transformation layers.
    
\begin{figure}[ht]
        \centering
        \includegraphics[width=0.7\textwidth]{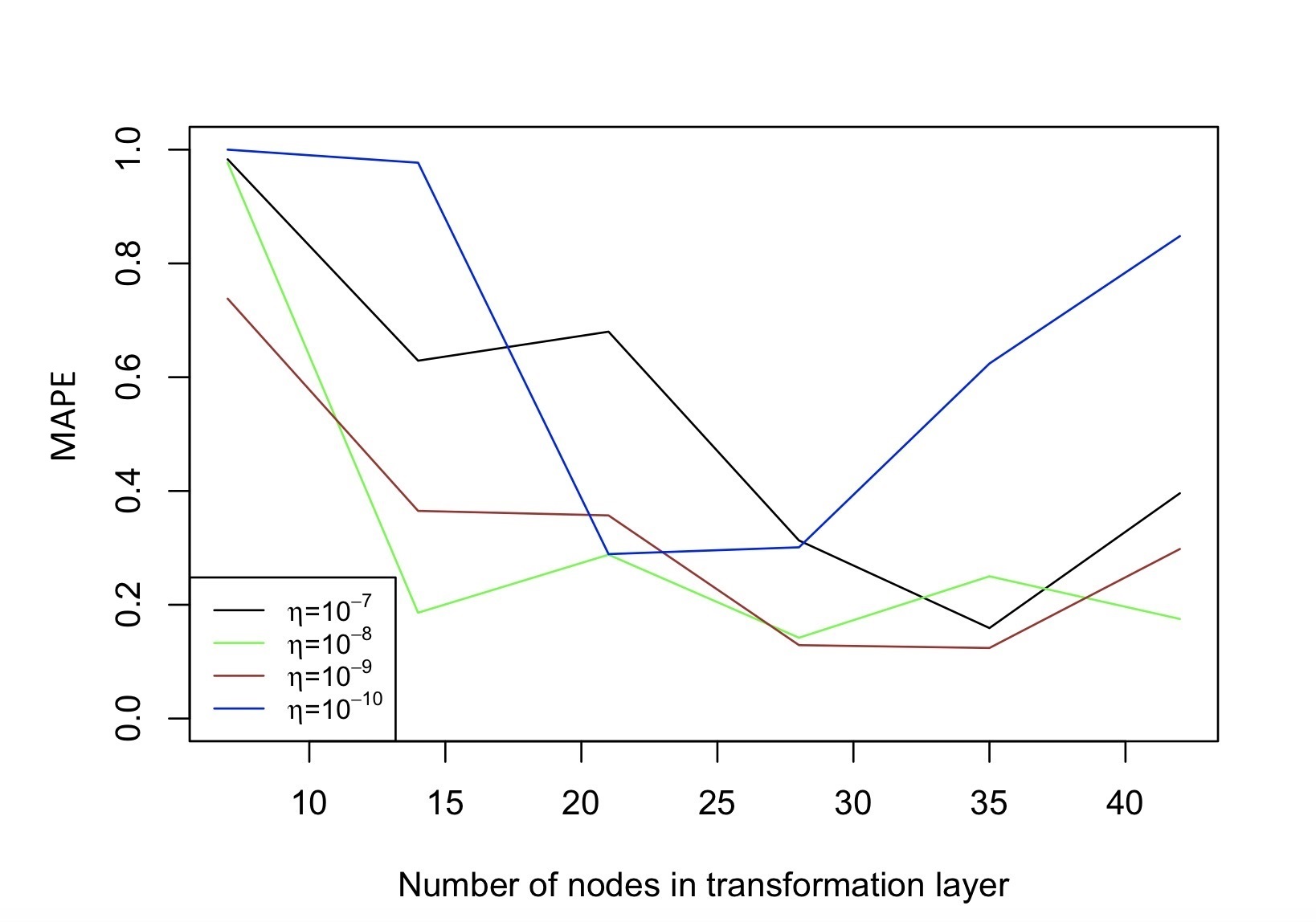}
        \caption{MAPE under different learning rates and size of nodes in transformation layers}
        \label{fig:RMSE_for_different_rate_and_nodes}
    \end{figure}

    Figure \ref{fig:RMSE_for_different_rate_and_nodes} shows the MAPE of PPGPR under different learning rate with respect to the size of representation nodes. It can be seen that when $\eta=10^{-10}$, the MAPE is much higher than those in the other three situations. For $\eta=10^{-8}$, the model reaches its best performance when $M=28$. The models with $M=35$ have lower MAPE when $\eta=10^{-7}$ and $\eta=10^{-9}$. In general,  the models with $\eta=10^{-9}$ perform slightly better and more stably.
    
    According to \cite{friedman2001elements}, the PPGPR model can approximate any continuous functions as $M\rightarrow\infty$ for an appropriate choice of kernel function. The Mat\'ern kernels are within this class because the reproducing kernel Hilbert space generated by any Mat\'ern kernel contains all polynomials. This explains why the performance of PPGPR grows as $M$ increases when $M$ is small. However, when $M$ is above $35$, the MAPE becomes worse for most of the curves in Figure \ref{fig:RMSE_for_different_rate_and_nodes}, which may be due to overfitting because there are too many hidden nodes. 
    In practice, we suggest employing cross-validation to select the optimal $M$.
    
\begin{figure}[ht]
        \centering
        \includegraphics[width=0.8\textwidth]{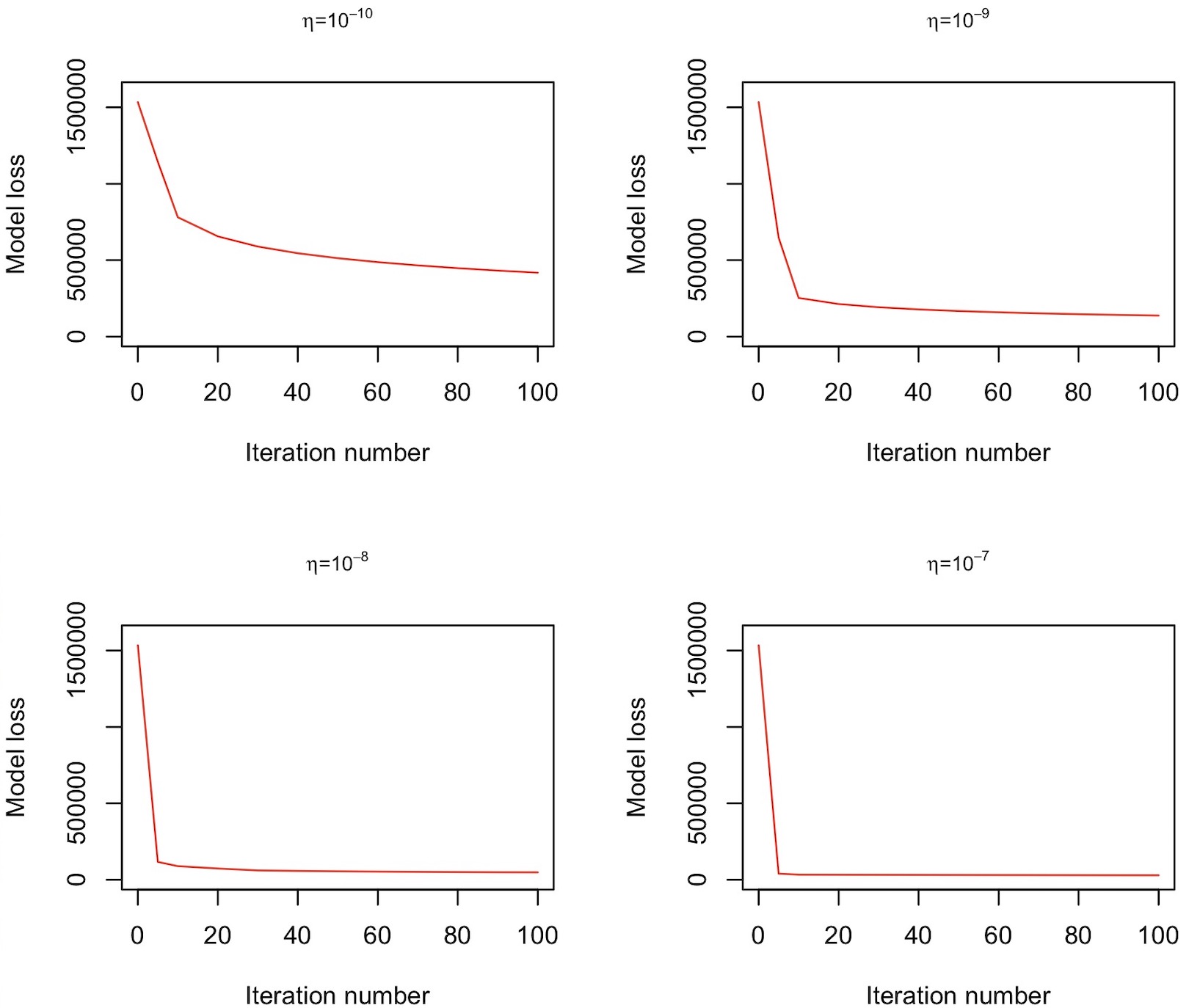}
        \caption{Model loss with different learning rate}
        \label{fig:different_rate}
    \end{figure}

    Figure \ref{fig:different_rate} shows four curves generated with a common initial $w$ and different learning rates when $M=35$. Each of them shows the relationship between the model loss defined in (\ref{objective function}) and the number of iteration. 
    From Figure \ref{fig:different_rate}, we find that, $10^{-10}$ is too low as a learning rate, because the model loss is still high (about $5\times 10^{5}$) even after 100 iterations. This observation is also confirmed by the MAPE results in Figure \ref{fig:RMSE_for_different_rate_and_nodes}, in which the MAPE for $M=35$ corresponding to $\eta=10^{-10}$ is much higher than those in the other ones. The model loss curves for the other three learning rates are similar. We believe that the choice of $\eta=10^{-9}$ gives a slightly better result than those given by $\eta=10^{-8}$ or $\eta=10^{-7}$, because the model loss curve under $\eta=10^{-9}$ decreases more smoothly than the other two, which implies a more stable learning process \citep{lawrence2000overfitting}. According to \cite{keskar2016large}, a flat minima might have higher generalization than sharp minima. Besides, a too small model loss after training might result in overfitting which will be shown in Section \ref{Sec:training epochs}. Figure \ref{fig:RMSE_for_different_rate_and_nodes} also implies that $\eta=10^{-9}$ gives the best MAPE when $M=35$.
    In practice, the optimal learning rate relies on the underlying function. Therefore, we recommend tuning $\eta$ via cross-validation.

    \subsubsection{Effects of correlation function type and parameters}\label{Sec:kernel function}
    In this experiment we examine the performance of PPGPR under different correlation functions and smoothness parameters with $\eta=10^{-9}$ and $P=150$.

\begin{figure}[ht]
        \centering
        \includegraphics[width=0.7\textwidth]{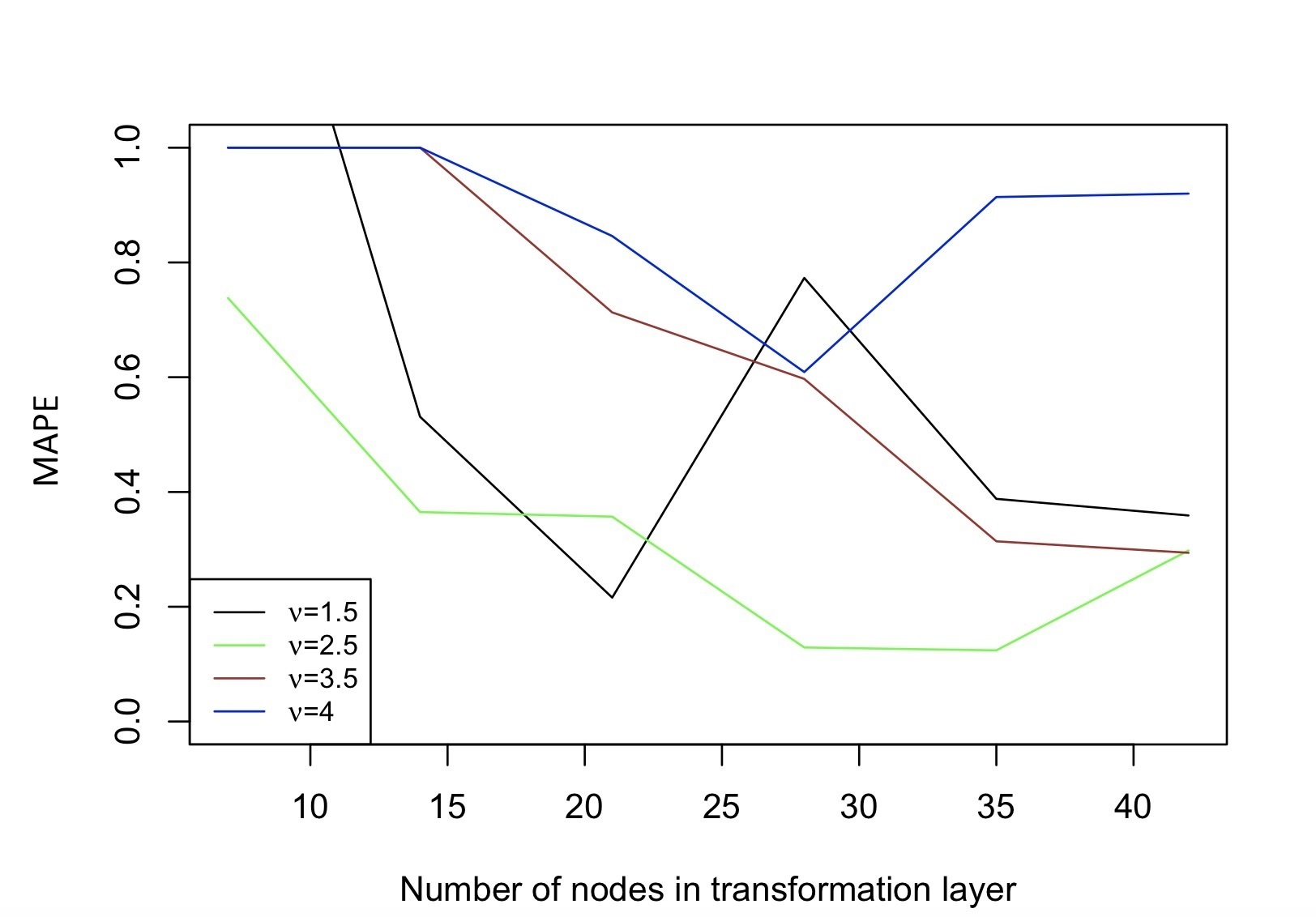}
        \caption{MAPE under different $\nu$ and $M$ for Mat\'ern correlation functions}
        \label{fig:Matern}
    \end{figure}

\begin{figure}[ht]
        \centering
        \includegraphics[width=0.7\textwidth]{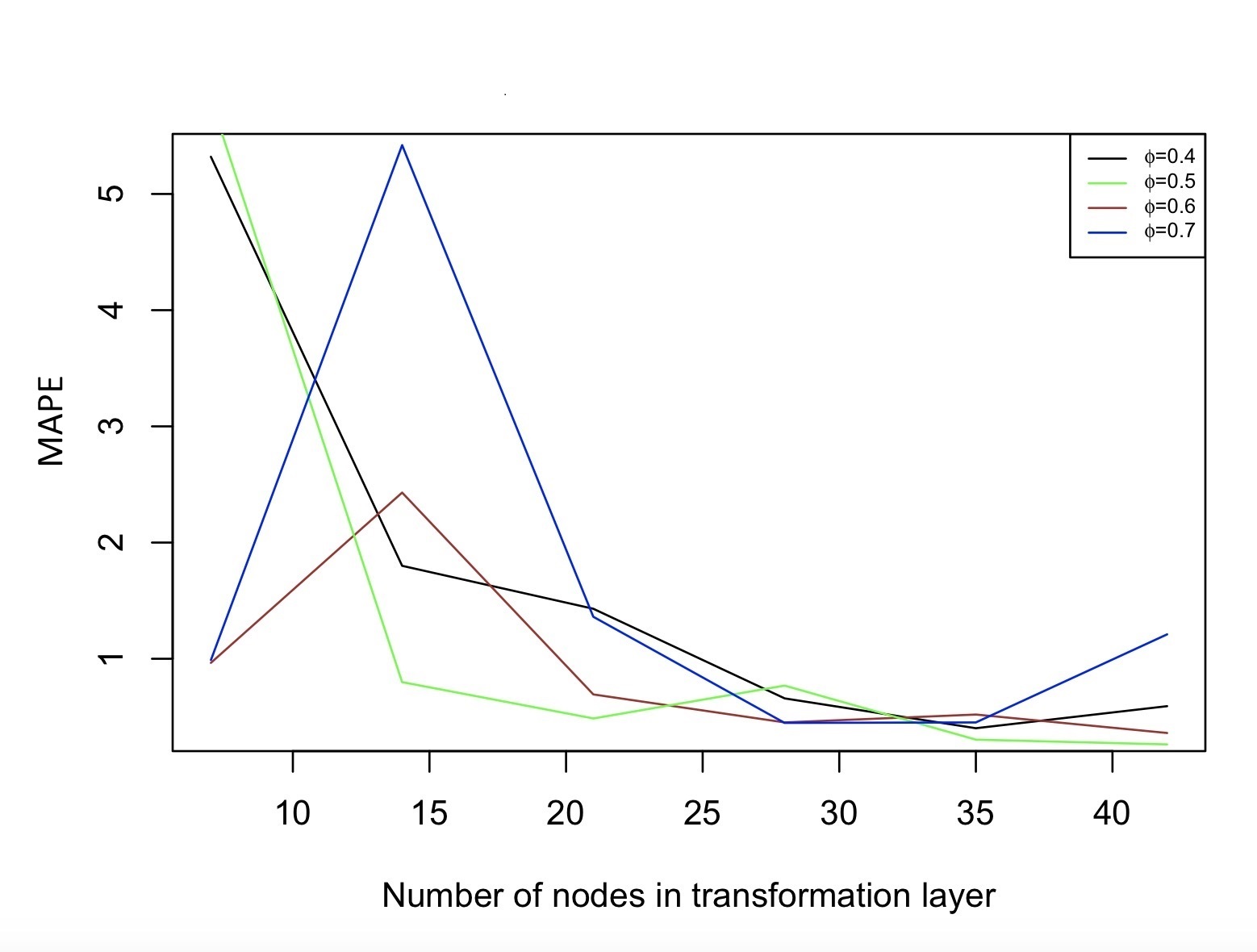}
        \caption{MAPE under different $\phi$ and $M$ for Gaussian correlation functions}
        \label{fig:gaussian}
    \end{figure}

    
    Figure \ref{fig:Matern} shows the MAPE for PPGPR with the Mat\'ern correlation functions under different $M$ and $\nu$ with $\phi=1$. It can be seen that when $\nu=2.5$ (green line), the model performs better than other choices. Under $\nu=2.5$, the best prediction performance is achieved when $M=35$. Generally, with a larger $\nu$, the reconstructed function would be smoother, which may lead to overfitting; with a smaller $\nu$, the reconstructed function would be less smooth, which may result in instability or underfitting. Figure \ref{fig:gaussian} shows the MAPE for PPGPR with Gaussian correlation functions under different $M$ and $\phi$. We can see that, when $M=35$, the green line ($\phi=0.5$) reaches its lowest MAPE, which is slightly better than the MAPE under other $M$ and $\phi$ in this experiment. This experiment shows that Mat\'ern correlation functions with $\nu=2.5$ seem to be an appropriate choice of the correlation functions. We also recommend using cross-validation to determine the optimal correlation function if computational resource permits. Table \ref{tab:table1} shows the numerical values of the lowest MAPE of PPGPR under the above Mat\'ern and Gaussian correlation functions. 

        {
    \begin{table}[ht]

        \caption{Best MAPE for PPGPR with Gaussian and Mat\'ern correlation functions}
        \begin{center}
        \begin{tabular}{ccccc}\hline
        \toprule
             & $M$ & $\phi$ & $\nu$ & MAPE  \\
        \midrule
             Mat\'ern & 35 & 1 & 2.5 & 0.124 \\
        \midrule
             Gaussian & 42 & 0.5 & - & 0.263 \\
        \bottomrule
        \end{tabular}
        \end{center}
        \label{tab:table1}
        
    \end{table}
    
    }
    
    \subsubsection{Training epochs $P$}\label{Sec:training epochs}
    In this experiment the model loss and the prediction error of PPGPR during the training process are monitored. Here we use a Mat\'ern correlation function with $\nu=2.5$ and $\eta=10^{-8}$, $M=21$.
    
\begin{figure}[tbhp]
        \centering
        \subfloat[Model loss in training process]{\label{fig:model_loss}\includegraphics[width=0.45\linewidth]{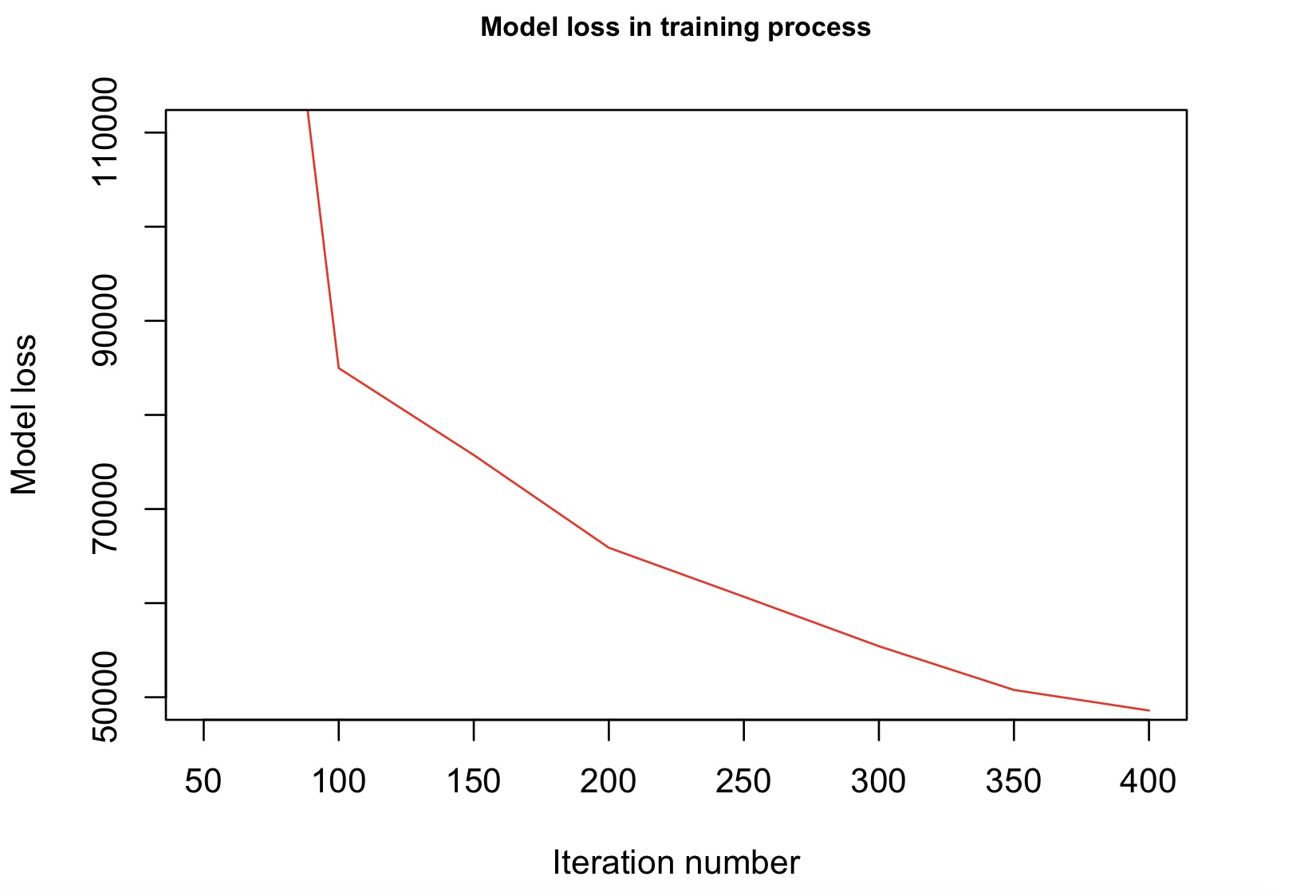}}
        \subfloat[Precision during the training process]{\label{fig:prediction_error}\includegraphics[width=0.45\linewidth]{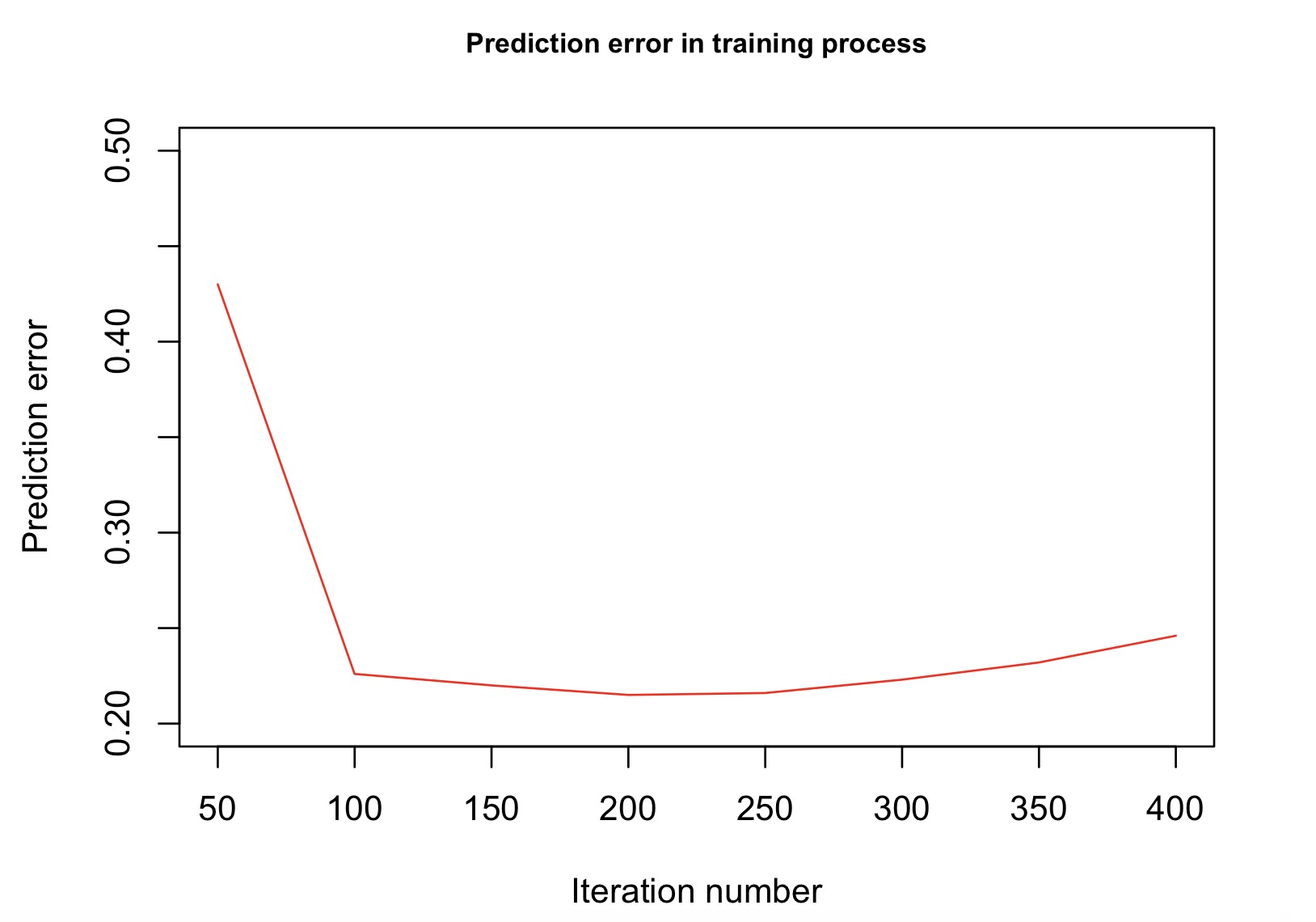}}
        \caption{Model loss and precision during the training process}
        \label{fig:loss_and_precision}
    \end{figure}

    Figures \ref{fig:model_loss} and \ref{fig:prediction_error} plot the model loss and prediction error against the training epochs, respectively. We can see from Figure \ref{fig:model_loss} that the model loss is monotonically decreasing as $M$ increases. This implies that the proposed gradient descent algorithm works in a desired way. However, Figure \ref{fig:prediction_error} shows that the prediction error is \textit{not} a monotonic function in the model loss. The model achieves its best performance when $P=220$, and as $P$ further increases, the prediction error increases. This phenomenon has been observed in other network structures such as neural networks. 
    In a typical neural network training process, a slower early-stopping criterion with $4\%$ (i.e., stopping the training process when the relative generalization improvement is less than $4\%$) could be used to avoid overfitting caused by an overshot training  process \citep{prechelt1998early}. We suggest adopting a similar approach in training the proposed PPGPR model.
    
    \subsection{Numerical comparisons}\label{Sec:numerical comparisons}
    In this section we compare PPGPR with GPR, Neural Network (NN), SVR (Supporting Vector Regression) and GBDT (Gradient Boosting Decision Trees) using three test functions: OTL circuit function \citep{ben2007modeling}, Borehole function \citep{harper1983sensitivity}, Wingweight function \citep{forrester2008engineering} and Welch function \citep{welch1992screening}. The training set is chosen as Halton series \citep{halton1964algorithm} with length $p=5 \times d$, where $d$ is the dimension of input space, and the size of testing set is 500. The implementation details of five methods for these experiments are shown below:
    \begin{itemize}
        \item SVR: Mat\'ern correlation with $\nu=2.5$;
        \item GBDT: Gaussian distribution and 100 trees;
        \item NN (deep learning): for the OTL circuit function, it has structure $(6,12,24,12,1)$ (meaning the node size of input layer is $6$, the second layer has 12 nodes and so on) with learning rate 0.01 and $150$ epochs. For the Borehole function, it has structure $(8,16,32,1)$ with learning rate $0.01$ and 150 epochs. For the Wingweight function, it has structure $(10,20,30,20,1)$ with learning rate $0.1$ and 200 epochs. For the Welch function, it has structure $(10,20,30,20,1)$ with learning rate $0.1$ and 200 epochs;
        \item GPR (with isotropic and product correlation functions): We use the Dicekriging package \citep{roustant2012dicekriging} with isotropic and product Mat\'ern correlation and smoothness $\nu=2.5$ to compute the predictive results. The product correlation is defined as $K(x)=\prod_{i=1}^{n}\Phi_1(x_{(j)})$, where $\Phi_1(x_{(j)})$ is the same as in (\ref{additive});
        \item PPGPR: for OTL circuit function, Mat\'ern correlation with $\nu=2.5$, $M=42$, $\eta=10^{-9}$, $P=150$, for Borehole function, Mat\'ern correlation with $\nu=2.5$, $M=35$, $\eta=10^{-9}$, $P=150$, for Wingweight function, Mat\'ern correlation with $\nu=2.5$, $M=35$, $\eta=10^{-10}$, $P=150$, for Welch function, Mat\'ern correlation with $\nu=2.5$, $M=7$, $\eta=10^{-6}$, $P=200$.
    \end{itemize}
    
    

    The MAPE of each method above is given in Table \ref{tab:table2}. It can be seen that the performances of SVR and GBDT are inferior in most cases, which can be explained because these approaches may require more training data \citep{ke2017lightgbm,smola2004tutorial}. The only exception is the case of the Welch function, where the SVR and the PPGPR result in comparable results. We have tried our best to tune the parameters of the NN, in order to obtain the best achievable results. It is worth noting that the parameter tuning for NN is time-consuming. In contrast, the tuning process of PPGPR is much easier because it has only one hidden layer. Also, PPGPR outperforms NN in all three experiments. Moreover, PPGPR can beat GPR with isotropic and product correlation functions because the curse of dimensionality has less impact on PPGPR. Note that GPR with isotropic kernels performs worse than GPR with product kernels. This is not surprising in view of the slow rate of convergence for isotropic kernels shown in Section 2.1. The rate of convergence for product kernels under a general condition is not well-established, but they are known to outperform the isotropic kernels in high-dimensional circumstances \citep{sacks1989design}.
    
    \begin{table}[ht]
        \centering
        \caption{MAPEs of Supporting Vector Regression (SVR), Gradient Boosting Decision Trees (GBDT), Neural Network (NN), Gaussian Process Regression (GPR) with isotropic and product correlations and Projection Pursuit Gaussian Process Regression (PPGPR) for three functions. PPGPR outperforms all other methods.}
        \begin{center}
        \begin{tabular}{ccccc}\hline
        \toprule
             & OTL circuit($d=6$) & Borehole($d=8$) &  Wingweight($d=10$) &Welch($d=20$)\\
        \midrule
             SVR & 0.121 & 0.792 & 0.127 & 0.989\\
        \midrule
             GBDT & 0.130 & 0.407 & 0.142 & 1.778\\
        \midrule
             NN & 0.0334 & 0.222 & 0.240&1.113\\
        \midrule
             GPR(iso) & 0.0182 & 0.204 & 0.0224 &1.334 \\
        \midrule
             GPR(pro) & 0.0162 & 0.134 & 0.0199 & 1.058 \\
        \midrule
             \textbf{PPGPR} & \bf 0.0139 & \bf 0.124 & \bf 0.0184 & \bf 0.994\\
        \bottomrule
        \end{tabular}
        
        \label{tab:table2}
        \end{center}
    \end{table}

\begin{figure}[ht]
        \centering
        \includegraphics[width=0.7\textwidth]{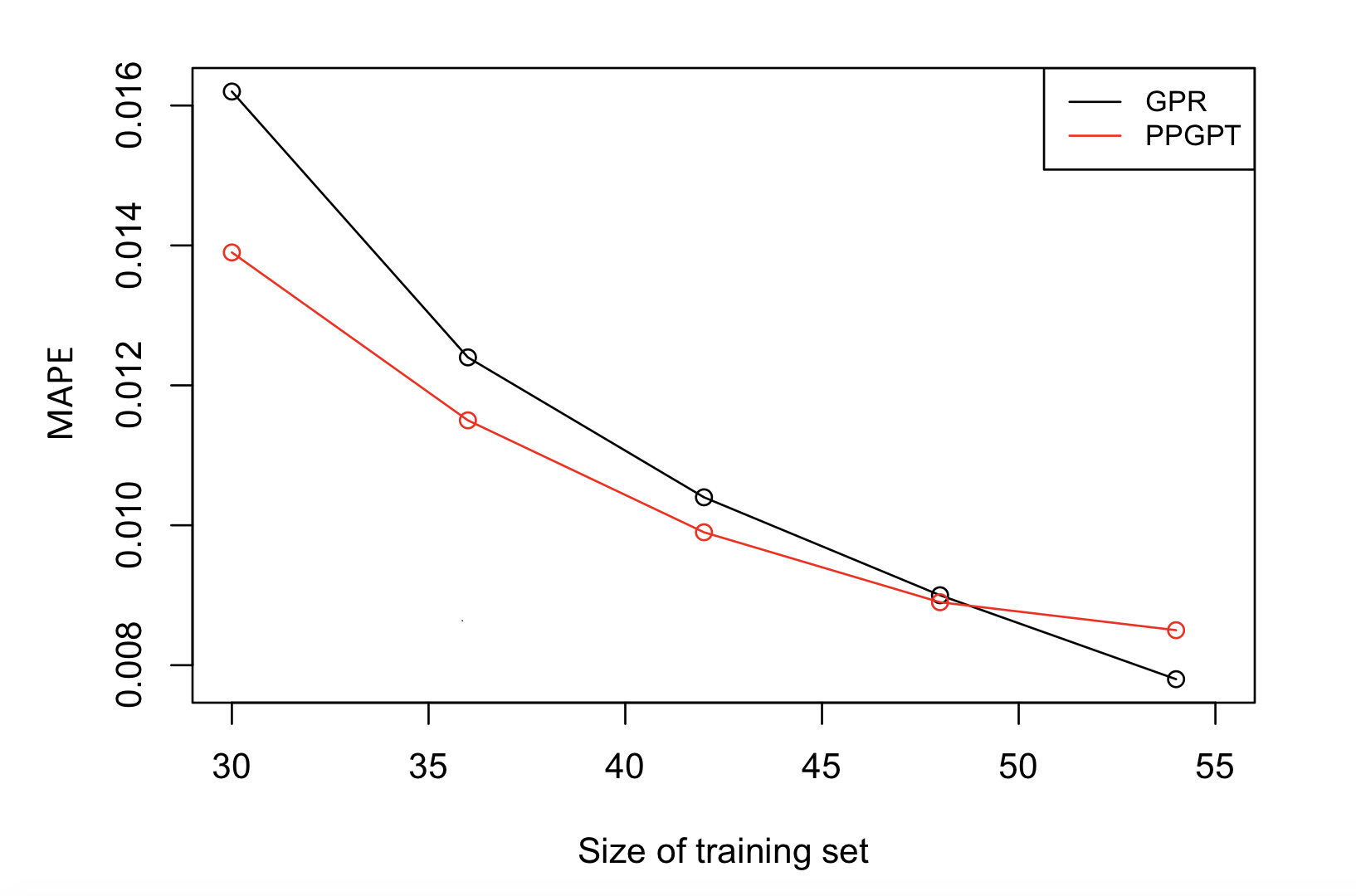}
        \caption{MAPEs of GPR and PPGPR with different size of training set for OTL function}
        \label{fig:different_size}
    \end{figure}

    Additionally, we compare the performance of PPGPR and GPR with product kernel when the size of training set changes. Figure \ref{fig:different_size} shows the MAPEs of the proposed PPGPR and GPR with product kernels for OTL function, when the number of training set varies. It can be seen that when the size of training samples is less than 48 ($8d$) the PPGPR works much better than GPR. When the size of training set increases the MAPEs of both methods decrease and the MAPE of GPR decreases faster than PPGPR. The results in Figure \ref{fig:different_size} can prove that the proposed PPGPR is highly suitable for a sparse learning environment but when enough traning samples are available the PPGPR is not recommended.  

    \subsection{Performance of GPR and PPGPR under Latin hypercube designs with different sizes}
\begin{figure}[ht]
        \centering
        \includegraphics[width=0.8\textwidth]{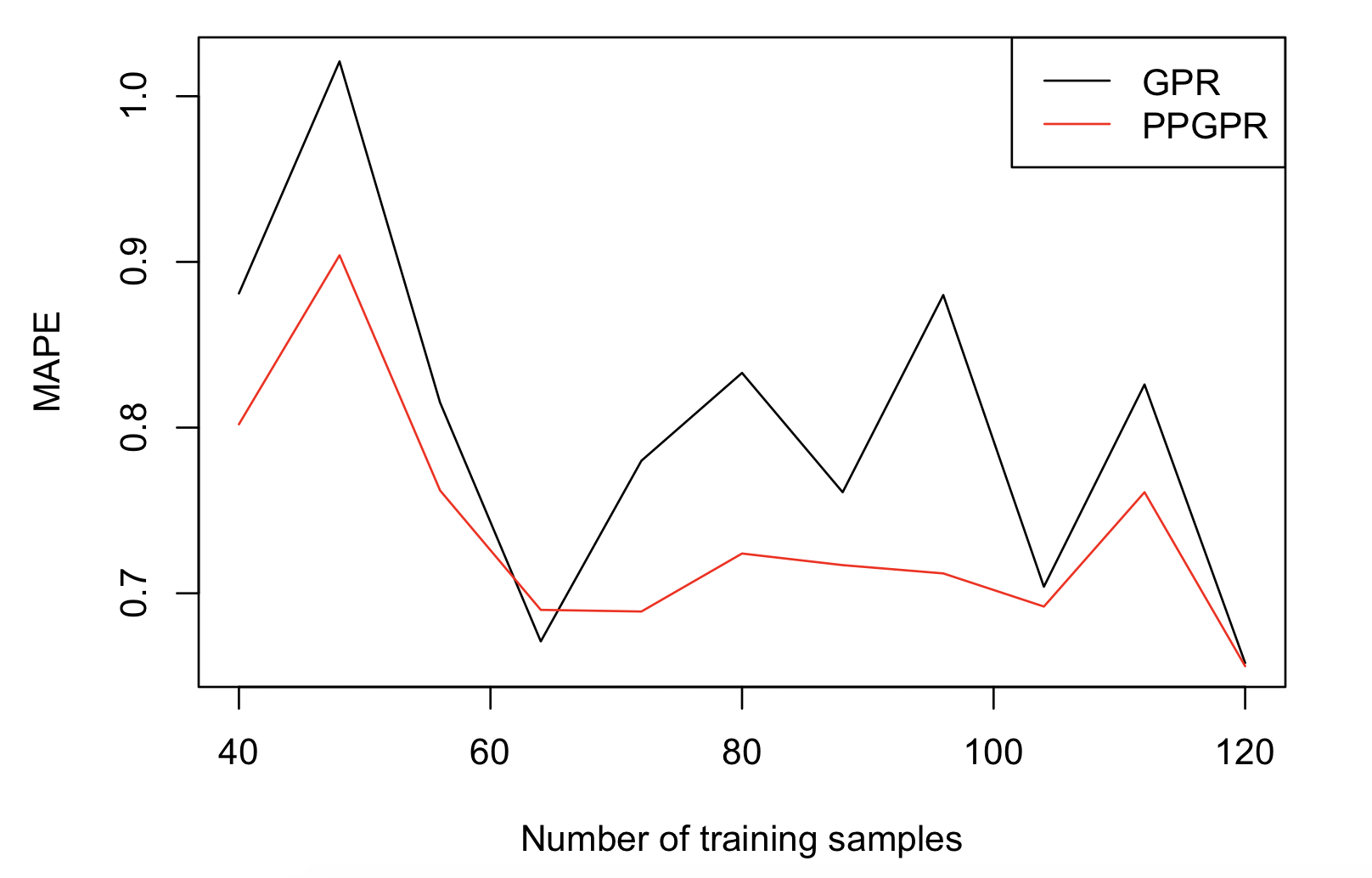}
        \caption{MAPEs of PPGPR and GPR for Dette Pepelyshev (2010) curved function under Latin hypercube designs with different sample sizes.}
        \label{fig:different_size2}
        \end{figure}

    We compare the performance of GPR and PPGPR under Latin hypercube designs \citep{helton2003latin} with different sample sizes. We choose the Dette Pepelyshev (2010) curved function \citep{dette2010generalized} as the underlying function. The  \texttt{R} package \texttt{lhs} 
    is used to generate the Latin hypercube designs using the maximin criterion \citep{joseph2008orthogonal}. The size of the testing set is $500$. Figure \ref{fig:different_size2} shows the MAPEs of GPR and PPGPR under the sample sizes from $40$ to $120$. It can be easily seen that the PPGPR has lower MAPEs than GPR most time. GPR has a lower MAPE only when the sample size is 63. Figure \ref{fig:different_size2} proves the superiority of the proposed PPGPR over GPR under the Latin hypercube design with different sample sizes. 

\section{More numerical studies}\label{more studies}
    We conduct more numerical studies to examine the computational cost of the proposed method, and the effect of initial values of the weight $w$. We also compare the PPGPR with a new additive Gaussian model proposed in \cite{delbridge2020randomly}. We defer these results to the Supplementary Materials.

\section{{Approximated heat exchanger case study}}\label{application}
    In this section, we apply the proposed method PPGPR on a heat exchanger (HE) application introduced by \cite{qian2006building}. The HE data in \cite{qian2006building} have two fidelities, known as detailed data (high fidelity) and approximated data (low fidelity). Because this work considers only the surrogate modeling for single-fidelity datasets, we use only the approximated data to implement the proposed method. The main objective of this application is to explore the impact of four factors, including the mass flow rate of entry air $m$, the temperature of entry air $T_{in}$, the temperature of the heat source $T_{wall}$ and the solid material thermal conductivity $M$, on the total rate of steady state heat transfer $y_a$ achieved by a heat exchanger. All design points live in a hypercube whose upper and lower bounds are shown in Table \ref{tab:table3}. We follow the treatment in \cite{qian2006building} to partition the dataset into a training set of 64 samples and a testing set of 14 samples. 
    \begin{table}[ht]
        
        \caption{Assumed design range for HE case}
        \begin{center}
        \begin{tabular}{ccccc}
        \toprule
             & $m(kg/s)$ & $T_{in}(K)$ & $k (W/mK)$ &  $T_{wall}(K)$ \\
        \midrule
             Lower Bound & 0.00055 & 270.00 & 202.4& 330\\
        \midrule
             Upper Bound & 0.001& 303.15& 360.0& 400 \\

        \bottomrule
        \end{tabular}
        
        \label{tab:table3}
        \end{center}
    \end{table}
    
    In this section, we compare the performance of GPR with isotropic and product correlations, Transformed Approximately Additive Gaussian Process Regression (TAAG) proposed in \cite{lin2019transformation}, and the proposed PPGPR. In \cite{lin2019transformation} the performance was assessed in terms of the root mean square error (RMSE), defined as
    \begin{eqnarray}\label{loss1}
        RMSE=\sqrt{\frac{1}{n}\sum_{i=1}^{n}(\hat{y}_i-y_i)^2},
    \end{eqnarray}
    where $\hat{y}_i$ is the predicted value and $y_i$ means the true value for every sample, $n$ stands for the size of testing set. Therefore, we  consider RMSE of all the candidate methods. The implementation details of the these methods are as follows:
    \begin{itemize}
        \item GPR (with isotropic and product correlation functions): we use the Dicekriging package \citep{roustant2012dicekriging} with isotropic and product Mat\'ern correlation and smoothness $\nu=2.5$ to compute the predictive results;
        \item TAAG: the result in \citep{lin2019transformation} is refered here;
        \item PPGPR: we use the Mat\'ern correlation with smoothness $\nu=2.5$ and $M=28$, $\eta = 10^{-9}$.
    \end{itemize}
    
    The results of these three methods are shown in Table \ref{tab:table4}. It can be seen that the proposed method has a lower RMSE than other methods.
        \begin{table}[ht]
        
        \caption{RMSEs of Gaussian Process Regression (GPR) with isotropic and product correlations, Transformed Approximately Additive Gaussian Process Regression (TAAG) and the Projection Pursuit Gaussian Process Regression (PPGPR)}
        \begin{center}
        \begin{tabular}{ccccc}\hline
        \toprule
             & GPR(iso) & GPR(pro) & TAAG &  PPGPR \\
        \midrule
             RMSE & 4.20 & 4.26 & 2.08 & 1.82\\

        \bottomrule
        \end{tabular}
        \end{center}
        \label{tab:table4}
    \end{table}

\section{Discussion}
\label{sec:discussion}
    In this paper, we propose a projection pursuit approach based on Gaussian process regression to fit deterministic computer outputs. The proposed method has a better model prediction and generalization power when the input dimension is high, and the sample size is small.

    Despite its advantages, the proposed method has a few issues to be addressed in future investigations.
    First, PPGPR involves quite a few hyper-parameters. Although we have provided a few guidelines regarding the choice of these hyper-parameters, how to better choose or tune these parameters requires further investigation. Second the current algorithm can only handle moderate data sets due to its high computational cost. We believe that this issue can be mitigated by implementing the following techniques: 1) parallel or GPU computation, 2) the recent advances in scalable GP inference and prediction \citep{liu2020gaussian,katzfuss2021general,chen2022kernel}.


    In practice, uncertainty quantification is often of importance in addition to a point estimation. Note that (\ref{prediction function}) can be regarded as an original GPR with an additive kernel function. In view of this, the corresponding confidence intervals can be obtained following a standard GPR technique. However, our numerical experience implies that, the confident bands provided by the above approach are much wider than those generated by the usual GPR methods. This deficiency may be due to the lack of identifiablity of the proposed models, as discussed in Section \ref{Sec:PPGPR}. Uncertainty quantification for the proposed model using alternative approaches should be considered in a future work.

\section*{Supplementary Materials}
In the Supplementary Materials, we present an upper bound of uniform prediction error of Gaussian process regression with an additive correlation function, which implies a promising rate of convergence of additive Gaussian process models. Also, more numerical studies are included in the Supplementary Materials. 

\section*{Acknowledgements}
The authors are grateful to the AE and two reviewers for very helpful comments and suggestions. This work is supported by NSF grants DSM-1914636 and CCF-1934904, and 2022 Texas A\&M Institute of Data Science Career Initiation Fellow Program.

\bibliographystyle{chicago}
\spacingset{1}
\bibliography{IISE-Trans}

\begin{thebibliography}{}

\bibitem[\protect\citeauthoryear{Ba and Joseph}{Ba and
  Joseph}{2012}]{ba2012composite}
Ba, S. and Joseph (2012).
\newblock Composite {Gaussian} process models for emulating expensive
  functions.
\newblock {\em The Annals of Applied Statistics\/}~{\em 6\/}(4), 1838--1860.

\bibitem[\protect\citeauthoryear{Ben-Ari and Steinberg}{Ben-Ari and
  Steinberg}{2007}]{ben2007modeling}
Ben-Ari, E.~N. and D.~M. Steinberg (2007).
\newblock Modeling data from computer experiments: an empirical comparison of
  kriging with mars and projection pursuit regression.
\newblock {\em Quality Engineering\/}~{\em 19\/}(4), 327--338.

\bibitem[\protect\citeauthoryear{Chen, Ding, and Tuo}{Chen
  et~al.}{2022}]{chen2022kernel}
Chen, H., L.~Ding, and R.~Tuo (2022).
\newblock Kernel packet: An exact and scalable algorithm for {G}aussian process
  regression with {M}at\'ern correlations.
\newblock {\em Journal of Machine Learning Research\/}~{\em 23\/}(127), 1--32.

\bibitem[\protect\citeauthoryear{Choi, Shi, and Wang}{Choi
  et~al.}{2011}]{choi2011gaussian}
Choi, T., J.~Q. Shi, and B.~Wang (2011).
\newblock A {Gaussian} process regression approach to a single-index model.
\newblock {\em Journal of Nonparametric Statistics\/}~{\em 23\/}(1), 21--36.

\bibitem[\protect\citeauthoryear{Constantine, Dow, and Wang}{Constantine
  et~al.}{2014}]{constantine2014active}
Constantine, P.~G., E.~Dow, and Q.~Wang (2014).
\newblock Active subspace methods in theory and practice: applications to
  kriging surfaces.
\newblock {\em SIAM Journal on Scientific Computing\/}~{\em 36\/}(4),
  A1500--A1524.

\bibitem[\protect\citeauthoryear{Delbridge, Bindel, and Wilson}{Delbridge
  et~al.}{2020}]{delbridge2020randomly}
Delbridge, I., D.~Bindel, and A.~G. Wilson (2020).
\newblock Randomly projected additive gaussian processes for regression.
\newblock In {\em International Conference on Machine Learning}, pp.\
  2453--2463. PMLR.

\bibitem[\protect\citeauthoryear{Deng, Lin, Liu, and Rowe}{Deng
  et~al.}{2017}]{deng2017additive}
Deng, X., C.~D. Lin, K.-W. Liu, and R.~Rowe (2017).
\newblock Additive {Gaussian} process for computer models with qualitative and
  quantitative factors.
\newblock {\em Technometrics\/}~{\em 59\/}(3), 283--292.

\bibitem[\protect\citeauthoryear{Dette and Pepelyshev}{Dette and
  Pepelyshev}{2010}]{dette2010generalized}
Dette, H. and A.~Pepelyshev (2010).
\newblock Generalized latin hypercube design for computer experiments.
\newblock {\em Technometrics\/}~{\em 52\/}(4), 421--429.

\bibitem[\protect\citeauthoryear{Durrande, Ginsbourger, and Roustant}{Durrande
  et~al.}{2012}]{durrande2012additive}
Durrande, N., D.~Ginsbourger, and O.~Roustant (2012).
\newblock Additive covariance kernels for high-dimensional gaussian process
  modeling.
\newblock In {\em Annales de la Facult{\'e} des sciences de Toulouse:
  Math{\'e}matiques}, Volume~21, pp.\  481--499.

\bibitem[\protect\citeauthoryear{Durrande, Ginsbourger, Roustant, and
  Carraro}{Durrande et~al.}{2013}]{durrande2013anova}
Durrande, N., D.~Ginsbourger, O.~Roustant, and L.~Carraro (2013).
\newblock Anova kernels and rkhs of zero mean functions for model-based
  sensitivity analysis.
\newblock {\em Journal of Multivariate Analysis\/}~{\em 115}, 57--67.

\bibitem[\protect\citeauthoryear{Duvenaud, Nickisch, and Rasmussen}{Duvenaud
  et~al.}{2011}]{duvenaud2011additive}
Duvenaud, D.~K., H.~Nickisch, and C.~E. Rasmussen (2011).
\newblock Additive {Gaussian} processes.
\newblock In {\em Advances in Neural Information Processing Systems}, pp.\
  226--234.

\bibitem[\protect\citeauthoryear{Ferraty, Goia, Salinelli, and Vieu}{Ferraty
  et~al.}{2013}]{ferraty2013functional}
Ferraty, F., A.~Goia, E.~Salinelli, and P.~Vieu (2013).
\newblock Functional projection pursuit regression.
\newblock {\em Test\/}~{\em 22\/}(2), 293--320.

\bibitem[\protect\citeauthoryear{Forrester, Sobester, and Keane}{Forrester
  et~al.}{2008}]{forrester2008engineering}
Forrester, A., A.~Sobester, and A.~Keane (2008).
\newblock {\em Engineering Design via Surrogate Modelling: A Practical Guide}.
\newblock John Wiley \& Sons.

\bibitem[\protect\citeauthoryear{Friedman, Hastie, and Tibshirani}{Friedman
  et~al.}{2001}]{friedman2001elements}
Friedman, J., T.~Hastie, and R.~Tibshirani (2001).
\newblock {\em The Elements of Statistical Learning}, Volume~1.
\newblock Springer series in statistics New York.

\bibitem[\protect\citeauthoryear{Friedman and Stuetzle}{Friedman and
  Stuetzle}{1981}]{friedman1981projection}
Friedman, J.~H. and W.~Stuetzle (1981).
\newblock Projection pursuit regression.
\newblock {\em Journal of the American statistical Association\/}~{\em
  76\/}(376), 817--823.

\bibitem[\protect\citeauthoryear{Gilboa, Saat{\c{c}}i, and Cunningham}{Gilboa
  et~al.}{2013}]{gilboa2013scaling}
Gilboa, E., Y.~Saat{\c{c}}i, and J.~Cunningham (2013).
\newblock Scaling multidimensional {G}aussian processes using projected
  additive approximations.
\newblock In {\em International Conference on Machine Learning}, pp.\
  454--461.

\bibitem[\protect\citeauthoryear{Glaws, Constantine, and Cook}{Glaws
  et~al.}{2020}]{glaws2020inverse}
Glaws, A., P.~G. Constantine, and R.~D. Cook (2020).
\newblock Inverse regression for ridge recovery: a data-driven approach for
  parameter reduction in computer experiments.
\newblock {\em Statistics and Computing\/}~{\em 30\/}(2), 237--253.

\bibitem[\protect\citeauthoryear{Goodfellow, Bengio, and Courville}{Goodfellow
  et~al.}{2016}]{goodfellow2016deep}
Goodfellow, I., Y.~Bengio, and A.~Courville (2016).
\newblock {\em Deep Learning}.
\newblock MIT press.

\bibitem[\protect\citeauthoryear{Gramacy and Lee}{Gramacy and
  Lee}{2008}]{gramacy2008bayesian}
Gramacy, R.~B. and H.~K.~H. Lee (2008).
\newblock Bayesian treed {G}aussian process models with an application to
  computer modeling.
\newblock {\em Journal of the American Statistical Association\/}~{\em
  103\/}(483), 1119--1130.

\bibitem[\protect\citeauthoryear{Gramacy and Lian}{Gramacy and
  Lian}{2012}]{gramacy2012gaussian}
Gramacy, R.~B. and H.~Lian (2012).
\newblock Gaussian process single-index models as emulators for computer
  experiments.
\newblock {\em Technometrics\/}~{\em 54\/}(1), 30--41.

\bibitem[\protect\citeauthoryear{Gu}{Gu}{2019}]{gu2019jointly}
Gu, M. (2019).
\newblock Jointly robust prior for gaussian stochastic process in emulation,
  calibration and variable selection.
\newblock {\em Bayesian Analysis\/}~{\em 14\/}(3), 857--885.

\bibitem[\protect\citeauthoryear{Halton}{Halton}{1964}]{halton1964algorithm}
Halton, J.~H. (1964).
\newblock Algorithm 247: Radical-inverse quasi-random point sequence.
\newblock {\em Communications of the ACM\/}~{\em 7\/}(12), 701--702.

\bibitem[\protect\citeauthoryear{Harper and Gupta}{Harper and
  Gupta}{1983}]{harper1983sensitivity}
Harper, W.~V. and S.~K. Gupta (1983).
\newblock {\em Sensitivity/Uncertainty Analysis of A Borehole Scenario
  Comparing Latin Hypercube Sampling and Deterministic Sensitivity Approaches}.
\newblock Office of Nuclear Waste Isolation, Battelle Memorial Institute.

\bibitem[\protect\citeauthoryear{Heaton, Christensen, and Terres}{Heaton
  et~al.}{2017}]{heaton2017nonstationary}
Heaton, M.~J., W.~F. Christensen, and M.~A. Terres (2017).
\newblock Nonstationary {Gaussian} process models using spatial hierarchical
  clustering from finite differences.
\newblock {\em Technometrics\/}~{\em 59\/}(1), 93--101.

\bibitem[\protect\citeauthoryear{Helton and Davis}{Helton and
  Davis}{2003}]{helton2003latin}
Helton, J.~C. and F.~J. Davis (2003).
\newblock Latin hypercube sampling and the propagation of uncertainty in
  analyses of complex systems.
\newblock {\em Reliability Engineering \& System Safety\/}~{\em 81\/}(1),
  23--69.

\bibitem[\protect\citeauthoryear{Hinton and Salakhutdinov}{Hinton and
  Salakhutdinov}{2006}]{hinton2006reducing}
Hinton, G.~E. and R.~R. Salakhutdinov (2006).
\newblock Reducing the dimensionality of data with neural networks.
\newblock {\em Science\/}~{\em 313\/}(5786), 504--507.

\bibitem[\protect\citeauthoryear{Hokanson and Constantine}{Hokanson and
  Constantine}{2018}]{hokanson2018data}
Hokanson, J.~M. and P.~G. Constantine (2018).
\newblock Data-driven polynomial ridge approximation using variable projection.
\newblock {\em SIAM Journal on Scientific Computing\/}~{\em 40\/}(3),
  A1566--A1589.

\bibitem[\protect\citeauthoryear{Hu, Gramacy, and Lian}{Hu
  et~al.}{2013}]{hu2013bayesian}
Hu, Y., R.~B. Gramacy, and H.~Lian (2013).
\newblock Bayesian quantile regression for single-index models.
\newblock {\em Statistics and Computing\/}~{\em 23\/}(4), 437--454.

\bibitem[\protect\citeauthoryear{James and Silverman}{James and
  Silverman}{2005}]{james2005functional}
James, G.~M. and B.~W. Silverman (2005).
\newblock Functional adaptive model estimation.
\newblock {\em Journal of the American Statistical Association\/}~{\em
  100\/}(470), 565--576.

\bibitem[\protect\citeauthoryear{Joseph and Hung}{Joseph and
  Hung}{2008}]{joseph2008orthogonal}
Joseph, V.~R. and Y.~Hung (2008).
\newblock Orthogonal-maximin latin hypercube designs.
\newblock {\em Statistica Sinica\/}, 171--186.

\bibitem[\protect\citeauthoryear{Katzfuss and Guinness}{Katzfuss and
  Guinness}{2021}]{katzfuss2021general}
Katzfuss, M. and J.~Guinness (2021).
\newblock A general framework for vecchia approximations of gaussian processes.
\newblock {\em Statistical Science\/}~{\em 36\/}(1), 124--141.

\bibitem[\protect\citeauthoryear{Ke, Meng, Finley, Wang, Chen, Ma, Ye, and
  Liu}{Ke et~al.}{2017}]{ke2017lightgbm}
Ke, G., Q.~Meng, T.~Finley, T.~Wang, W.~Chen, W.~Ma, Q.~Ye, and T.-Y. Liu
  (2017).
\newblock Lightgbm: A highly efficient gradient boosting decision tree.
\newblock In {\em Advances in Neural Information Processing Systems}, pp.\
  3146--3154.

\bibitem[\protect\citeauthoryear{Keskar, Mudigere, Nocedal, Smelyanskiy, and
  Tang}{Keskar et~al.}{2016}]{keskar2016large}
Keskar, N.~S., D.~Mudigere, J.~Nocedal, M.~Smelyanskiy, and P.~T.~P. Tang
  (2016).
\newblock On large-batch training for deep learning: Generalization gap and
  sharp minima.
\newblock {\em arXiv preprint arXiv:1609.04836\/}.

\bibitem[\protect\citeauthoryear{Khoo, Lu, and Ying}{Khoo
  et~al.}{2017}]{khoo2017solving}
Khoo, Y., J.~Lu, and L.~Ying (2017).
\newblock Solving parametric {PDE} problems with artificial neural networks.
\newblock {\em arXiv preprint arXiv:1707.03351\/}.

\bibitem[\protect\citeauthoryear{Lawrence and Giles}{Lawrence and
  Giles}{2000}]{lawrence2000overfitting}
Lawrence, S. and C.~L. Giles (2000).
\newblock Overfitting and neural networks: conjugate gradient and
  backpropagation.
\newblock In {\em Proceedings of the IEEE-INNS-ENNS International Joint
  Conference on Neural Networks. IJCNN 2000. Neural Computing: New Challenges
  and Perspectives for the New Millennium}, Volume~1, pp.\  114--119. IEEE.

\bibitem[\protect\citeauthoryear{Lebarbier}{Lebarbier}{2005}]{lebarbier2005detecting}
Lebarbier, {\'E}. (2005).
\newblock Detecting multiple change-points in the mean of {Gaussian} process by
  model selection.
\newblock {\em Signal Processing\/}~{\em 85\/}(4), 717--736.

\bibitem[\protect\citeauthoryear{LeCun, Bengio, and Hinton}{LeCun
  et~al.}{2015}]{lecun2015deep}
LeCun, Y., Y.~Bengio, and G.~Hinton (2015).
\newblock Deep learning.
\newblock {\em Nature\/}~{\em 521\/}(7553), 436--444.

\bibitem[\protect\citeauthoryear{Li, Kandasamy, P{\'o}czos, and Schneider}{Li
  et~al.}{2016}]{li2016high}
Li, C.-L., K.~Kandasamy, B.~P{\'o}czos, and J.~Schneider (2016).
\newblock High dimensional {Bayesian} optimization via restricted projection
  pursuit models.
\newblock In {\em Artificial Intelligence and Statistics}, pp.\  884--892.

\bibitem[\protect\citeauthoryear{Lin and Roshan~Joseph}{Lin and
  Roshan~Joseph}{2020}]{lin2019transformation}
Lin, L.-H. and V.~Roshan~Joseph (2020).
\newblock Transformation and additivity in {Gaussian} processes.
\newblock {\em Technometrics\/}~{\em 62\/}(4), 525--535.

\bibitem[\protect\citeauthoryear{Linkletter, Bingham, Hengartner, Higdon, and
  Ye}{Linkletter et~al.}{2006}]{linkletter2006variable}
Linkletter, C., D.~Bingham, N.~Hengartner, D.~Higdon, and K.~Q. Ye (2006).
\newblock Variable selection for {Gaussian} process models in computer
  experiments.
\newblock {\em Technometrics\/}~{\em 48\/}(4), 478--490.

\bibitem[\protect\citeauthoryear{Liu, Ong, Shen, and Cai}{Liu
  et~al.}{2020}]{liu2020gaussian}
Liu, H., Y.-S. Ong, X.~Shen, and J.~Cai (2020).
\newblock When gaussian process meets big data: A review of scalable gps.
\newblock {\em IEEE transactions on neural networks and learning
  systems\/}~{\em 31\/}(11), 4405--4423.

\bibitem[\protect\citeauthoryear{Mak, Sung, Wang, Yeh, Chang, Joseph, Yang, and
  Wu}{Mak et~al.}{2018}]{mak2018efficient}
Mak, S., C.-L. Sung, X.~Wang, S.-T. Yeh, Y.-H. Chang, V.~R. Joseph, V.~Yang,
  and C.~F.~J. Wu (2018).
\newblock An efficient surrogate model for emulation and physics extraction of
  large eddy simulations.
\newblock {\em Journal of the American Statistical Association\/}~{\em
  113\/}(524), 1443--1456.

\bibitem[\protect\citeauthoryear{Makridakis}{Makridakis}{1993}]{makridakis1993accuracy}
Makridakis, S. (1993).
\newblock Accuracy measures: theoretical and practical concerns.
\newblock {\em International journal of forecasting\/}~{\em 9\/}(4), 527--529.

\bibitem[\protect\citeauthoryear{M{\"u}ller and Yao}{M{\"u}ller and
  Yao}{2008}]{muller2008functional}
M{\"u}ller, H.-G. and F.~Yao (2008).
\newblock Functional additive models.
\newblock {\em Journal of the American Statistical Association\/}~{\em
  103\/}(484), 1534--1544.

\bibitem[\protect\citeauthoryear{Oakley and O'Hagan}{Oakley and
  O'Hagan}{2004}]{oakley2004probabilistic}
Oakley, J.~E. and A.~O'Hagan (2004).
\newblock Probabilistic sensitivity analysis of complex models: a {Bayesian}
  approach.
\newblock {\em Journal of the Royal Statistical Society: Series B (Statistical
  Methodology)\/}~{\em 66\/}(3), 751--769.

\bibitem[\protect\citeauthoryear{Peng and Wu}{Peng and
  Wu}{2014}]{peng2014choice}
Peng, C.-Y. and C.~F.~J. Wu (2014).
\newblock On the choice of nugget in kriging modeling for deterministic
  computer experiments.
\newblock {\em Journal of Computational and Graphical Statistics\/}~{\em
  23\/}(1), 151--168.

\bibitem[\protect\citeauthoryear{Pinkus}{Pinkus}{1997}]{pinkus1997approximating}
Pinkus, A. (1997).
\newblock Approximating by ridge functions.
\newblock {\em Surface fitting and multiresolution methods\/}, 279--292.

\bibitem[\protect\citeauthoryear{Prechelt}{Prechelt}{1998}]{prechelt1998early}
Prechelt, L. (1998).
\newblock Early stopping--but when?
\newblock In {\em Neural Networks: Tricks of the Trade}, pp.\  55--69.
  Springer.

\bibitem[\protect\citeauthoryear{Psichogios and Ungar}{Psichogios and
  Ungar}{1992}]{psichogios1992hybrid}
Psichogios, D.~C. and L.~H. Ungar (1992).
\newblock A hybrid neural network-first principles approach to process
  modeling.
\newblock {\em AIChE Journal\/}~{\em 38\/}(10), 1499--1511.

\bibitem[\protect\citeauthoryear{Qian, Seepersad, Joseph, Allen, and
  Jeff~Wu}{Qian et~al.}{2006}]{qian2006building}
Qian, Z., C.~C. Seepersad, V.~R. Joseph, J.~K. Allen, and C.~Jeff~Wu (2006).
\newblock Building surrogate models based on detailed and approximate
  simulations.

\bibitem[\protect\citeauthoryear{Rasmussen and Williams}{Rasmussen and
  Williams}{2006}]{Rasmussen2006gaussian}
Rasmussen, C. and C.~Williams (2006).
\newblock {\em Gaussian Processes for Machine Learning}.
\newblock The MIT Press.

\bibitem[\protect\citeauthoryear{Roustant, Ginsbourger, and Deville}{Roustant
  et~al.}{2012}]{roustant2012dicekriging}
Roustant, O., D.~Ginsbourger, and Y.~Deville (2012).
\newblock Dicekriging, diceoptim: Two {R} packages for the analysis of computer
  experiments by kriging-based metamodeling and optimization.

\bibitem[\protect\citeauthoryear{Sacks, Welch, Mitchell, and Wynn}{Sacks
  et~al.}{1989}]{sacks1989design}
Sacks, J., W.~Welch, T.~Mitchell, and H.~Wynn (1989).
\newblock Design and analysis of computer experiments.
\newblock {\em Statistical Science\/}~{\em 4\/}(4), 409--423.

\bibitem[\protect\citeauthoryear{Saltelli, Annoni, Azzini, Campolongo, Ratto,
  and Tarantola}{Saltelli et~al.}{2010}]{saltelli2010variance}
Saltelli, A., P.~Annoni, I.~Azzini, F.~Campolongo, M.~Ratto, and S.~Tarantola
  (2010).
\newblock Variance based sensitivity analysis of model output. design and
  estimator for the total sensitivity index.
\newblock {\em Computer Physics Communications\/}~{\em 181\/}(2), 259--270.

\bibitem[\protect\citeauthoryear{Santner, Williams, and Notz}{Santner
  et~al.}{2019}]{santner2003design}
Santner, T., B.~Williams, and W.~Notz (2019).
\newblock {\em The Design and Analysis of Computer Experiments}.
\newblock Springer; 2nd edition.

\bibitem[\protect\citeauthoryear{Smola and Sch{\"o}lkopf}{Smola and
  Sch{\"o}lkopf}{2004}]{smola2004tutorial}
Smola, A.~J. and B.~Sch{\"o}lkopf (2004).
\newblock A tutorial on support vector regression.
\newblock {\em Statistics and Computing\/}~{\em 14\/}(3), 199--222.

\bibitem[\protect\citeauthoryear{Stein}{Stein}{1999}]{stein1999interpolation}
Stein, M. (1999).
\newblock {\em Interpolation of Spatial Data: Some Theory for Kriging}.
\newblock Springer Verlag.

\bibitem[\protect\citeauthoryear{Tripathy, Bilionis, and Gonzalez}{Tripathy
  et~al.}{2016}]{tripathy2016gaussian}
Tripathy, R., I.~Bilionis, and M.~Gonzalez (2016).
\newblock Gaussian processes with built-in dimensionality reduction:
  Applications to high-dimensional uncertainty propagation.
\newblock {\em Journal of Computational Physics\/}~{\em 321}, 191--223.

\bibitem[\protect\citeauthoryear{Tripathy and Bilionis}{Tripathy and
  Bilionis}{2018}]{tripathy2018deep}
Tripathy, R.~K. and I.~Bilionis (2018).
\newblock Deep {UQ}: Learning deep neural network surrogate models for high
  dimensional uncertainty quantification.
\newblock {\em Journal of Computational Physics\/}~{\em 375}, 565--588.

\bibitem[\protect\citeauthoryear{Tuo and Wang}{Tuo and
  Wang}{2020}]{tuo2020kriging}
Tuo, R. and W.~Wang (2020).
\newblock Kriging prediction with isotropic {M}at\'ern correlations: Robustness
  and experimental designs.
\newblock {\em J. Mach. Learn. Res.\/}~{\em 21}, 1--38.

\bibitem[\protect\citeauthoryear{Wang, Xue, Zhu, and Chong}{Wang
  et~al.}{2010}]{wang2010estimation}
Wang, J.-L., L.~Xue, L.~Zhu, and Y.~S. Chong (2010).
\newblock Estimation for a partial-linear single-index model.
\newblock {\em The Annals of statistics\/}~{\em 38\/}(1), 246--274.

\bibitem[\protect\citeauthoryear{Welch, Buck, Sacks, Wynn, Mitchell, and
  Morris}{Welch et~al.}{1992}]{welch1992screening}
Welch, W.~J., R.~J. Buck, J.~Sacks, H.~P. Wynn, T.~J. Mitchell, and M.~D.
  Morris (1992).
\newblock Screening, predicting, and computer experiments.
\newblock {\em Technometrics\/}~{\em 34\/}(1), 15--25.

\bibitem[\protect\citeauthoryear{Wendland}{Wendland}{2005}]{wendland2005scattered}
Wendland, H. (2005).
\newblock {\em Scattered Data Approximation}.
\newblock Cambridge University Press.

\bibitem[\protect\citeauthoryear{Wilson, Knowles, and Ghahramani}{Wilson
  et~al.}{2011}]{wilson2011gaussian}
Wilson, A.~G., D.~A. Knowles, and Z.~Ghahramani (2011).
\newblock Gaussian process regression networks.
\newblock {\em arXiv preprint arXiv:1110.4411\/}.

\end{thebibliography}
	
\end{document}